\documentclass[lettersize,journal]{IEEEtran}
\usepackage{amsmath,amsfonts}
\usepackage{algorithmic}
\usepackage{algorithm}
\usepackage{array}
\usepackage[caption=false,font=normalsize,labelfont=sf,textfont=sf]{subfig}
\usepackage{textcomp}
\usepackage{stfloats}
\usepackage{url}
\usepackage{verbatim}
\usepackage{graphicx}
\usepackage{hyperref}  
\hypersetup{colorlinks}

\usepackage{booktabs}
\usepackage{color}
\usepackage{multirow}
\usepackage{xcolor}
\usepackage{subcaption}
\usepackage{colortbl}
\usepackage{tabularx}
\usepackage{bbding}
\usepackage{pifont}

\usepackage[nocompress]{cite} 

\newcommand{\eg}{\textit{e.g.}}

\begin{document}

\title{Boost UAV-based Ojbect Detection via Scale-Invariant Feature Disentanglement and Adversarial Learning}


\author{Fan Liu, \IEEEmembership{Member, IEEE}, Liang Yao, \IEEEmembership{Graduate Student Member, IEEE}, Chuanyi Zhang, \IEEEmembership{Member, IEEE}, Ting Wu, Xinlei Zhang, Xiruo Jiang \IEEEmembership{Member, IEEE}, and Jun Zhou, \IEEEmembership{Senior Member, IEEE}
\thanks{Fan Liu and Liang Yao contributed equally to this work. Corresponding author: Chuanyi Zhang (20231104@hhu.edu.cn).}
\thanks{Fan Liu, Liang Yao, Ting Wu, and Xinlei Zhang are with the College of Computer Science and Software Engineering, Hohai University, Nanjing, 210098, China.}
\thanks{Chuanyi Zhang is with the College of Artificial Intelligence and Automation, Hohai University, Nanjing, 210098, China.}
\thanks{Xiruo Jiang is with the School of Computer Science and Engineering, Nanjing University of Science and Technology,  Nanjing, 210094, China.}
\thanks{Jun Zhou is with the School of Information and Communication Technology, Griffith University, Nathan, Queensland 4111, Australia.}
}

\markboth{Journal of \LaTeX\ Class Files,~Vol.~14, No.~8, August~2021}%
{Shell \MakeLowercase{\textit{et al.}}: A Sample Article Using IEEEtran.cls for IEEE Journals}


\maketitle

\begin{abstract}
Detecting objects from Unmanned Aerial Vehicles (UAV) is often hindered by a large number of small objects, resulting in low detection accuracy. To address this issue, mainstream approaches typically utilize multi-stage inferences. Despite their remarkable detecting accuracies, real-time efficiency is sacrificed, making them less practical to handle real applications.
To this end, we propose to improve the single-stage inference accuracy through learning scale-invariant features. Specifically, a Scale-Invariant Feature Disentangling module is designed to disentangle scale-related and scale-invariant features. Then an Adversarial Feature Learning scheme is employed to enhance disentanglement. Finally, scale-invariant features are leveraged for robust UAV-based object detection. Furthermore, we construct a multi-modal UAV object detection dataset, State-Air, which incorporates annotated UAV state parameters. 
We apply our approach to three lightweight detection frameworks on two benchmark datasets. Extensive experiments demonstrate that our approach can effectively improve model accuracy and achieve state-of-the-art (SoTA) performance on two datasets.
Our code and dataset will be publicly available once the paper is accepted. 
\end{abstract}


\section{Introduction}

\begin{figure}[t]
    \centering
    \includegraphics[width=0.99\linewidth]{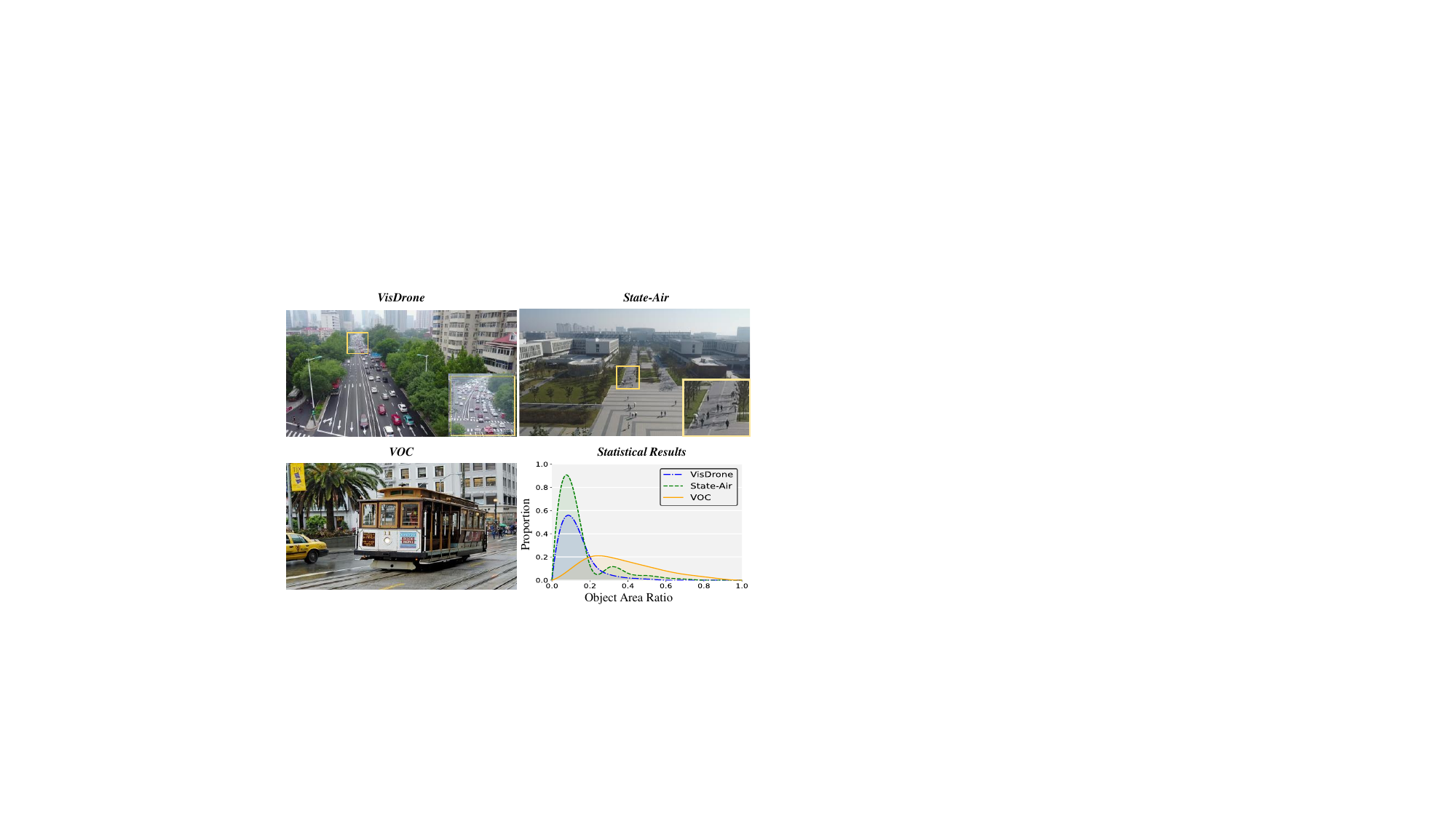}
    \caption{
    Comparison of general (VOC) and UAV (VisDrone, State-Air) datasets. The object scale is normalized by the ratio of the object's actual area to the source image. Proportion represents the percentage of objects of each scale in the overall dataset. Most objects in UAV datasets tend to be small-scale.}
    \label{fig1}
\end{figure}

With the rapid development of the Unmanned Aerial Vehicles (UAV) industry, UAV technology has been widely applied in various fields~\cite{huang2022object, luo2023evolutionary, khankeshizadeh2024novel,10476501} such as agriculture, logistics, and rescue~\cite{zhang2024empowering,cheng:uav-system,rejeb2022drones,srivastava2023techniques}. As one of the fundamental tasks in UAV applications, UAV-based object detection (UAV-OD) has attracted wide attention from the research community~\cite{su2023ai,mittal2020deep,wu2021deep,zitar2023review}.
As illustrated in Fig.~\ref{fig1}, a significant difference between general and UAV-based object detection is the viewing angle and the object scale.
Specifically, UAV tends to have a top-down view at a high altitude and most of the objects in the field of view are small-scale. 

Furthermore, the UAV altitude may change in the flight, resulting in a varying object scale~\cite{10098606}. If the extracted features are scale-related, they may interfere with robust detection of small objects.
In addition, UAV computing platforms tend to have limited computational capability~\cite{10745610}, they can hardly apply large object detectors~\cite{han2023typical,biswas2022improving,suh2023algorithm,ren2024dino}. Owing to the above drawbacks (small objects and restricted computational resources), UAV-based object detection is a challenging task.

To overcome the challenge of small and varying scale objects in UAV-OD, researchers typically employ multi-stage coarse-to-fine reasoning methods~\cite{xie2024fewer,huang2022ufpmp,duan2021coarse,li2021aerial}. 
Initially, a detector is utilized to roughly localize regions containing small objects. Subsequently, the area resolution is enlarged for refined small object detection. Although many of these methods have achieved remarkable accuracy, multiple inferences on a single image tend to be time-consuming. Therefore they are still not suitable for deployment on UAVs due to stringent real-time requirements.

Another approach to alleviating the problem of small object detection is to leverage scale-invariant features~\cite{park2023ssfpn,wang2020scale}. Scale-invariant features, such as shape features~\cite{lowe1999object}, remain unchanged regardless of variations in the object scale.
If a model can effectively learn scale-invariant features, the varying object scale issue can be mitigated and the capability to detect small objects can be enhanced.
However, during the process of extracting deep features, there is often an ambiguity between Scale-Related and Scale-Invariant features~\cite{van2017learning}. 
Consequently, it is difficult to extract specific scale-invariant features in an unsupervised manner. \cite{park2023ssfpn} proposed ssFPN to use a convolutional structure to extract scale-invariant features.
However, the capability of extracted features to handle the scale-invariant properties of objects in UAV-captured images is uncertain.

The widespread adoption of UAVs leads to the creation of a large number of UAV-OD datasets~\cite{zhu2018vision,du2018unmanned,robicquet2016learning}. 
While these datasets advance the UAV-OD, the majority of them neglect flight status data such as UAV-specific parameters and altitudes.
The flight status data can be potentially beneficial for UAV-OD research, and a minority of datasets capture this ancillary data, \eg, AU-AIR~\cite{bozcan2020air} and SynDrone~\cite{rizzoli2023syndrone}. Nevertheless, they tend to be plagued by issues such as imprecise annotations or a lack of diversity in environments.
In addition, as a simulated dataset, SynDrone tend to be less conducive to practical applications.


To this end, we propose a novel approach named \textbf{SIFDAL} (Scale-Invariant Feature Disentanglement via Adversarial Learning), a new plug-and-play module for effective single-stage UAV-OD. It is composed of a Scale-Invariant Feature Disentangling (\textbf{SIFD}) module and an Adversarial Feature Learning (\textbf{AFL}) training scheme.
Specifically, we first analyze the effect of various resolution layers of the feature pyramid network (FPN)~\cite{lin2017feature} on UAV-OD. The results indicate that the high-resolution layer of FPN plays a more vital role in small object detection.
Then the SIFD module is developed to disentangle scale-invariant features from the high-resolution feature map. 
Next, we utilize the AFL training scheme to realize the maximal disentanglement.
Finally, discriminative scale-invariant features can boost the detection accuracy.
Our SIFDAL can be easily extended to FPN-based object detectors (\eg, YOLOv7) to improve the accuracy. In addition, we propose a real-world multi-modal UAV-OD dataset named \textbf{State-Air}, which records UAV IMU (Inertial Measurement Unit) parameters and flight altitudes. 

Our contributions are summarized as follows:
\begin{itemize}
    \item We propose a scale-invariant feature disentangling module, which can be applied to any FPN-based object detector. To the best of our knowledge, this is the first method to improve UAV-based object detection accuracy by disentangling scale-invariant features.
    \item We introduce a training scheme with adversarial feature learning to enhance feature disentanglement. It significantly improves the disentanglement effect as well as detection accuracy.
    \item We construct a multi-scene and multi-modal UAV-based object detection dataset, \textbf{State-Air}. It incorporates UAV IMU parameters as well as flight altitudes and covers multiple scenes and weather conditions.
    \item We validate our proposed approach with extensive experiments on three UAV benchmarks by integrating SIFDAL into various base detectors with FPN. After employing our SIFDAL, YOLOv7 achieves state-of-the-art (SoTA) performance on three datasets.
\end{itemize}

The remainder of this paper is structured as follows. In Section~\ref{rw}, we review the relevant literature on UAV-based object detection, scale-invariant feature extraction, and UAV-OD Datasets. Section~\ref{method} introduces our SIFDAL. We first revisit FPN in UVA-OD in Section~\ref{m1}. Then we describe our scale-invariant feature disentangling method in Section~\ref{m2}. Section~\ref{m3} introduces an adversarial learning strategy to facilitate the disentangling process. In Section~\ref{Data}, we construct the State-Air dataset. Section~\ref{sec:experiment} presents our empirical evaluation of our SIFDAL. Finally, Section~\ref{discussion} presents our discussion of the advantages and limitations of our approach and concludes this paper.

\begin{figure*}[t]
    \centering
    \includegraphics[width=0.99\linewidth]{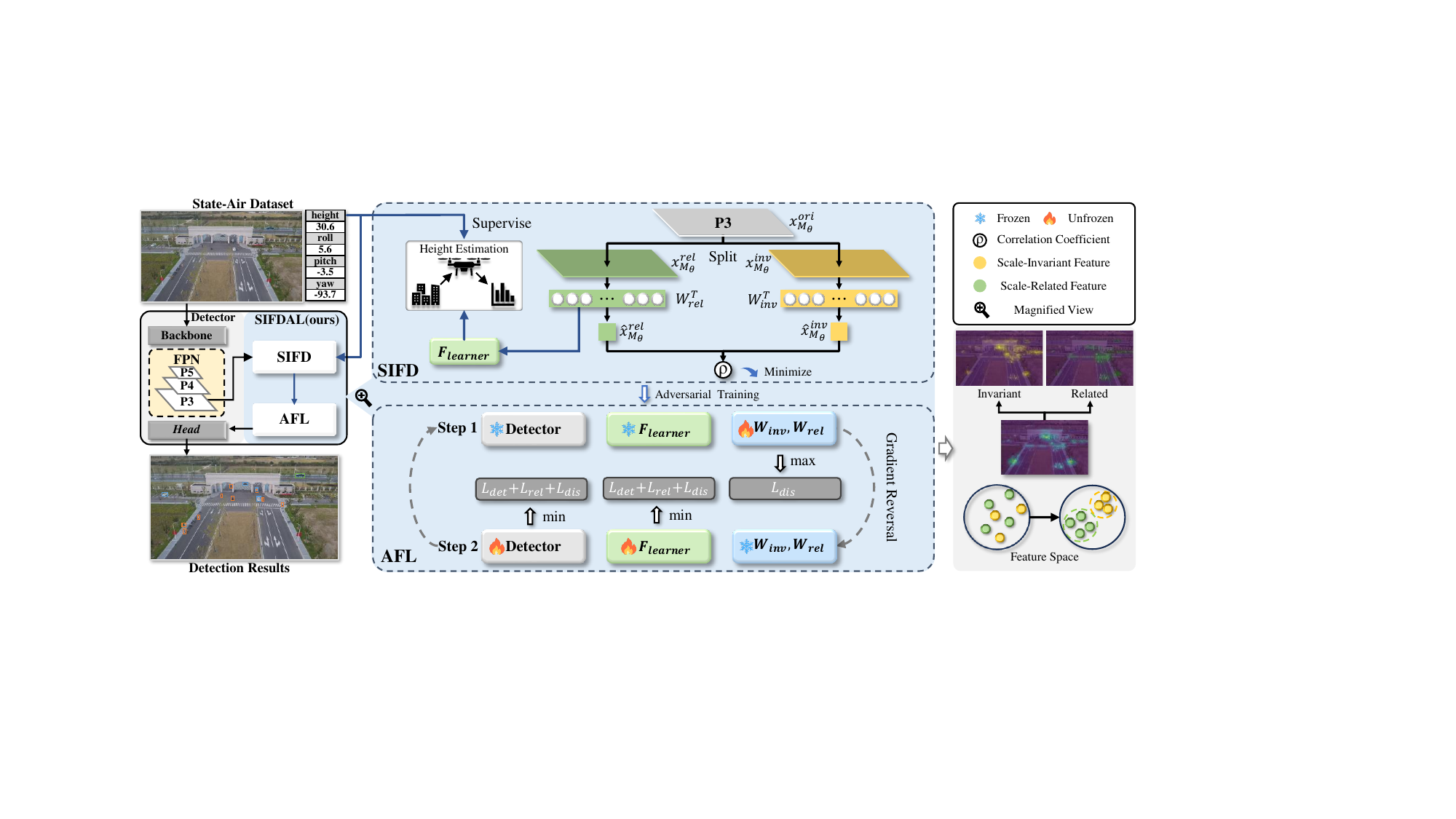}
    \caption{Overview of our proposed approach. State-Air is a multi-scene and multi-modal UAV-OD dataset that incorporates UAV state parameters. Our SIFDAL consists of a SIFD module and an AFL training method. SIFD leverages a scale-related feature learner $\mathcal{F}_{learner}$ to extract scale-related features through height level estimation with altitude labels as supervision. Then, it disentangles scale-related and scale-invariant features ($x_{\mathcal{M}_{\theta}}^{rel}$ and $x_{\mathcal{M}_{\theta}}^{inv}$) by minimizing the correlation coefficient $\rho$. AFL is utilized to enhance feature disentanglement by adversarial training. Finally, scale-invariant features are employed to detect objects.}
    \label{fig2}
\end{figure*}

\section{Related Work} \label{rw}
\subsection{UAV-based Object Detection}
Different from general object detection tasks~\cite{FasterRCNN,yolo,yolov4,lv2023detrs},
UAV-OD typically encounters the challenge of small objects and the limited computing power of edge equipment.
A typical solution to alleviate the small object detection problem is to adopt a coarse-to-fine strategy~\cite{duan2021coarse,li2020density,yang2019clustered,li2017perceptual}. Initially, large objects are detected, and small object-dense subregions are located. Then subregions are employed as model input to obtain further detection results.
In this step, a Gaussian mixture model could be used to supervise the detector in generating object clusters composed of focusing regions~\cite{koyun2022focus}.
Alternatively, a CZ detector~\cite{CZDetector} utilized a density crop labeling algorithm to label the crowded object regions and then upscaled those regions to augment the training data.

To alleviate the problem of restricted computing resources, some methods were proposed to balance the accuracy and the efficiency. For example, sparse convolution~\cite{liu2015sparse} was used to design lightweight network architectures that significantly reduce computing costs~\cite{figurnov2017spatially,yan2018second}.
Typical examples include Querydet~\cite{yang2022querydet} and CEASC~\cite{CEASC}. The former utilized a sparse detection head to enable fast and accurate small object detection. It introduced a novel query mechanism to accelerate the inference speed of FPN-based object detectors. 
The latter adopted a plug-and-play detection method with enhanced sparse convolution and an adaptive mask scheme.


\subsection{Scale-Invariant Feature Extraction}

Scale-invariant features remain unchanged even when the object scale varies, so it is widely used to address the multi-scale issue in computer vision.
For example, SIFT algorithm~\cite{SIFT} conducted Gaussian difference operations at various scales and directions to detect local key feature points. Its strong scale and rotation invariance property enabled effective image feature matching.
Later, SURF~\cite{SURF} was proposed to enhance SIFT with faster calculation and improved robustness.

A more common approach is to consider scale dependencies in feature pyramids.
For example, a Trident-FPN backbone network~\cite{lin2021ecascade} was designed to address the multi-scale problem in aerial images. It introduced a novel attention and anchor generation algorithm to enhance object detection performance.
\cite{wang2022near} employed scale-invariant features to transform visible and infrared images for time series alignment and matching.
\cite{behera2023superpixel} utilized super-pixel images with key context information to extract scale-invariant features for predicting the object class of each pixel. 
To prevent information loss of objects in deep structures, \cite{park2023ssfpn} proposed a scale-sequence-based feature extraction method for FPN. The FPN structure was viewed as a scale space, and the scale sequence features were extracted through 3D convolution as scale-invariant features.

\subsection{UAV-based Object Detection Datasets}
There are many datasets that support the development of drone-based object detection. For example, Stanford-Drone~\cite{Robicquet2016LearningSE} is a detailed dataset featuring overhead images and videos of various objects interacting at Stanford University. UAVDT~\cite{du2018unmanned} comprises 80,000 annotated frames with bounding boxes and 14 attributes across diverse scenarios.
VisDrone~\cite{du2019visdrone} is a large-scale benchmark for object detection and tracking, showcasing various environmental conditions and camera angles. It encompasses a total of 10 distinct categories of objects relevant to UAV applications, with over 2.5 million annotated bounding boxes.

With the development of multimodal learning, researchers are increasingly attempting to incorporate more modalities into datasets.
AU-AIR~\cite{bozcan2020air} provides frame metadata, bounding box annotations for traffic-related objects, and multi-modal flight sensor data, captured at low altitudes in various lighting conditions at intersections. SynDrone~\cite{rizzoli2023syndrone} features a multi-modal synthetic benchmark dataset with images and 3D data from different flying heights, including 28 classes with pixel-level labels and object-level annotations for semantic segmentation and object detection.

\section{Method} \label{method}

In this section, we introduce the proposed Scale-Invariant Feature Disentanglement via Adversarial Learning (SIFDAL) method. It consists of two components: Scale-Invariant Feature Disentangling (SIFD) and Adversarial Feature Learning (AFL). The overall framework of our approach with YOLOv7-L as the base detector is illustrated in Fig.~\ref{fig2} and and algorithm is demonstrated in Alg. \ref{alg1}.

\subsection{Revisiting FPN in UAV-OD} 
\label{m1}
When the convolutional neural network (CNN)~\cite{li2021survey} is applied for deep feature extraction, the network gradually decreases the resolution of the feature map. Consequently, small objects eventually vanish in deep layers. In most object detection methods, FPN is commonly utilized as the model's ``neck".
One of its functions is detecting objects of varying scales by entering features from each layer to the corresponding detection head. The detection head for a low-resolution layer is used to detect large objects, while one for a high-resolution layer is employed for small objects.

\begin{figure*}[t]
    \centering
    \includegraphics[width=0.98\linewidth]{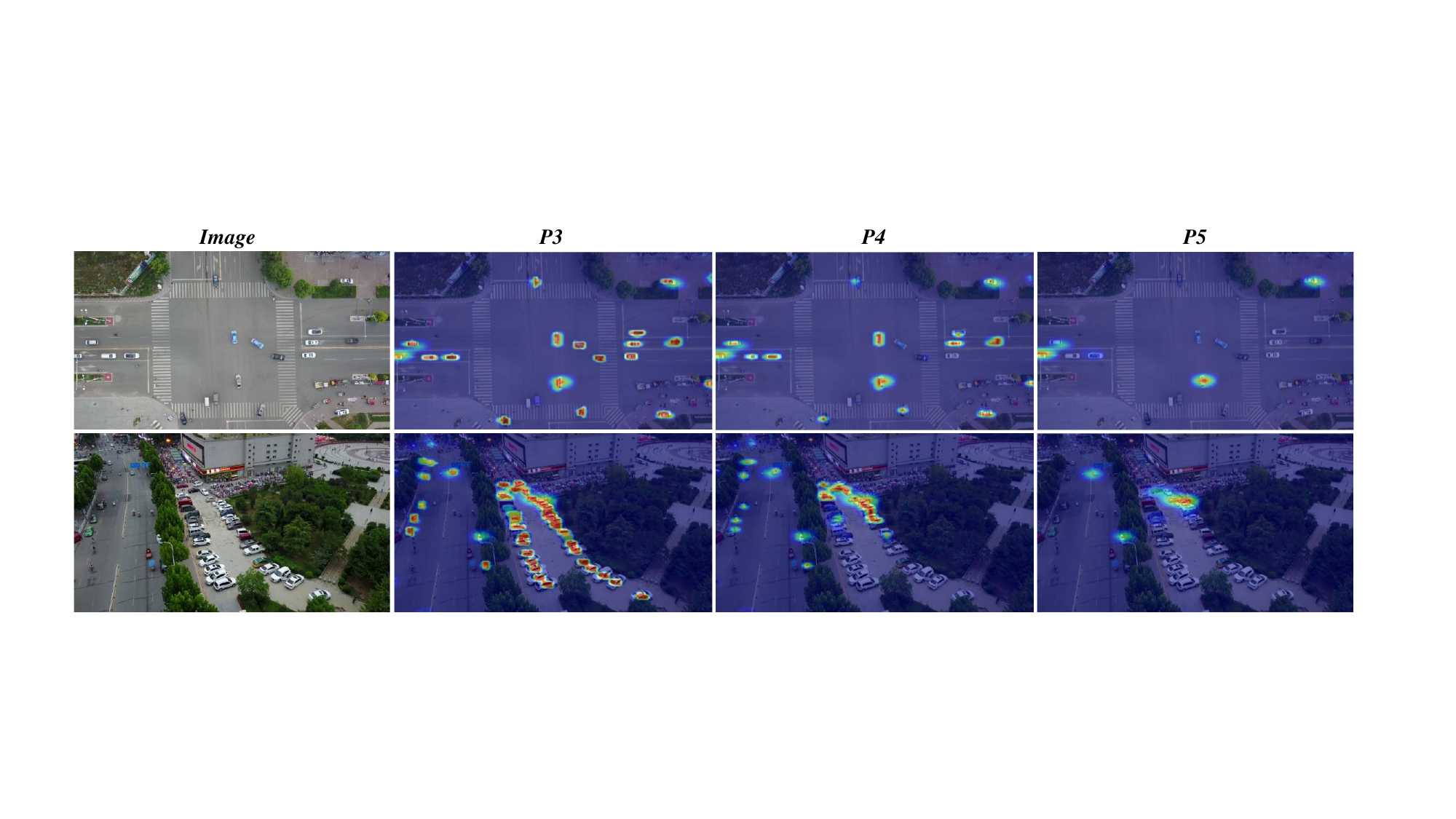}
    \caption{Visualization of heat maps in different FPN layers of YOLOv7-L. The resolution of the feature map decreases sequentially from P3 to P5. The detection head for the P3 layer corresponds to the majority of objects in the drone's field of view.}
    \label{fig3}
    \vspace{-0.2cm}
\end{figure*}

We visualize the heat map of different FPN layers in YOLOv7-L. As illustrated in Fig.~\ref{fig3}, the detection head with the high-resolution layer (P3) is responsible for the majority of objects in the UAV's visual field. With a large number of small objects, the high-resolution detection head frequently plays an important role in the UAV-based object detection tasks. To boost the accuracy of small object detection, we aim to guide the high-resolution detection head to leverage scale-invariant features via feature disentangling.


\subsection{Scale-Invariant Feature Disentangling} \label{m2}
After multi-scale feature fusion of FPN, object features often contain both scale-related and scale-invariant information. Since scale-invariant features are not easily affected by varying object scales, they can be more conducive to UAV-OD than scale-related ones. To enable the model to learn scale-invariant features, we design a SIFD module that can be utilized in any detection model with FPN. 

\subsubsection{Feature Splitting}
We directly apply channel splitting to the high-resolution layer in FPN, segmenting the feature map into two groups. The formula can be expressed as:
\begin{equation}
x_{\mathcal{M}_{\theta}}^{rel},x_{\mathcal{M}_{\theta}}^{inv} = Split(x_{\mathcal{M}_{\theta}}^{ori}),
\label{eq1}
\end{equation}
where $\mathcal{M}_{\theta}$ denotes the detector with parameters $\theta$, $x_{\mathcal{M}_{\theta}}^{ori} \in \mathbb{R}^{H \times W \times 2L}$ represents the original feature map extracted by $\mathcal{M}_{\theta}$, $x_{\mathcal{M}_{\theta}}^{rel} \in \mathbb{R}^{H \times W \times L}$ and $x_{\mathcal{M}_{\theta}}^{inv} \in \mathbb{R}^{H \times W \times L}$ are the two feature maps after splitting. $H$, $W$, and $L$ represent the height, width, and number of channels of the feature map, respectively.
Next, we will disentangle two features to make $x_{\mathcal{M}_{\theta}}^{rel}$ and $x_{\mathcal{M}_{\theta}}^{inv}$ learn scale-related and scale-invariant features, respectively. To accomplish this objective, we propose a scale-related loss and a feature disentangling loss.

\subsubsection{Scale-Related Loss}

\begin{algorithm*}[b]
    \caption{\textbf{Scale-Invariant Feature Disentanglement via Adversarial Learning (SIFDAL)}}
    \textbf{Require:} \\
        $\mathcal{M}_{\theta}$: Detector with parameter $\theta$, \\
        $x_{\mathcal{M}_{\theta}}^{ori} \in \mathbb{R}^{H \times W \times 2L}$: Feature map of $P_{3}$,\\
        $W=\{W_{rel},W_{inv}\}$: Linear layers,\\
        $\mathcal{L}_{det},\mathcal{L}_{rel},\mathcal{L}_{dis}$: Object detection loss, scale-related loss and feature disentangling loss.

    \begin{algorithmic}[1]
        \FOR{each training epoch}
            \FOR{each training sample feature 
            $x_{\mathcal{M}_{\theta}}^{ori}$}
                \STATE  \textit{\# Splitting feature map of $P_{3}$ into two parts.} \\                $x_{\mathcal{M}_{\theta}}^{rel},x_{\mathcal{M}_{\theta}}^{inv} = Split(x_{\mathcal{M}_{\theta}}^{ori})$ 
 \hfill Eq.~\eqref{eq1}\\ 
                \STATE
                \textit{\# Learning scale-related features through the height level estimation task.} \\
                $\mathcal{L}_{rel} \left(x_{\mathcal{M}_{\theta}}^{rel}, h; \theta, W_{h}^{T} \right)= \; - {{h \cdot {\mathit{\log} {\left( Softmax \left( W_{h}^{T} \cdot P_{avg}( x_{\mathcal{M}_{\theta}}^{rel})\right)\right)} }}}$ \hfill Eq.~\eqref{eq2}\\
                
                \STATE 
                \textit{\# Projecting $x_{\mathcal{M}_{\theta}}^{inv}$ and $x_{\mathcal{M}_{\theta}}^{rel}$ for calculating the correlation coefficient.} \\$\hat{x}_{\mathcal{M}_{\theta}}^{rel} = W_{rel}^{T} \cdot x_{\mathcal{M}_{\theta}}^{rel}$
                \STATE $\hat{x}_{\mathcal{M}_{\theta}}^{inv} = W_{inv}^{T}\cdot x_{\mathcal{M}_{\theta}}^{inv}$  \hfill Eq.~\eqref{eq3}\\

                \STATE $\mathcal{L}_{\rho} \left(x_{\mathcal{M}_{\theta}}^{rel}, x_{\mathcal{M}_{\theta}}^{inv}; \theta, W_{\mathit{rel}}, W_{\mathit{inv}}\right) = \rho^{2}\left( W_{rel}^{T} \cdot x_{\mathcal{M}_{\theta}}^{rel},W_{inv}^{T} \cdot x_{\mathcal{M}_{\theta}}^{inv} \right)$ \hfill  Eq.~\eqref{eq5}\\
                 \textit{\# The minimization process is assigned more iterations to accomplish feature disentanglement.} \\
                    \IF {in the first 30 of 80 iterations}
                        \STATE $Freeze(\mathcal{M_{\theta}}, W_{h}),\  Unfreeze(W_{\mathit{rel}}, W_{\mathit{inv}})$\\                      
                        \STATE $\mathcal{L}_{dis} \leftarrow \max\limits_{W}\mathcal{L}_{\rho}$  
                    \ELSE
                        \STATE $Unfreeze(W_{\mathit{rel}}, W_{\mathit{inv}}), \ Freeze(\mathcal{M_{\theta}}, W_{h})$\\
                        \STATE $\mathcal{L}_{dis} \leftarrow \min\limits_{\mathcal{M_{\theta}}}{\mathcal{L}}_{\rho}$  \hfill Eq.~\eqref{eq6}\\
                    \ENDIF
        
                \STATE $\mathcal{L} = \mathcal{L}_{det}  + ~\lambda_{1}\mathcal{L}_{rel} + {\lambda_{2}}\mathcal{L}_{dis}$
                \hfill Eq.~\eqref{eq8}\\
                \STATE Optimize $\mathcal{L}$ to find optimal $x_{\mathcal{M}_{\theta}}^{rel}$ and $x_{\mathcal{M}_{\theta}}^{inv}$
            \ENDFOR
        \ENDFOR
        \RETURN Disentangled features $x_{\mathcal{M}_{\theta}}^{rel}$ and $x_{\mathcal{M}_{\theta}}^{inv}$
    \end{algorithmic}
\label{alg1}
\end{algorithm*}

Intuitively, the object scale in the view is in connection with the UAV's altitude. As the UAV flies higher, the object scale becomes smaller. If $x_{\mathcal{M}_{\theta}}^{rel}$ is trained to correctly perform height estimation, it can be regarded as scale-related features. In other words, we can utilize the height estimation task for scale-related feature learning.

However, without relying on external knowledge such as laser ranging, it is difficult to predict specific heights solely from images. To reduce the task complexity, we replace the height regression with a simpler altitude classification task. Specifically, we accurately group the heights and label each image with a height level $h$. 
The height grouping is accomplished by the $k$-\textit{means} clustering algorithm~\cite{duda1973pattern}, where the group number $k$ is determined according to the characteristics of each dataset.
Finally, we can simply leverage a fully connected layer $W_{h}$ with a softmax operation as our scale-related feature learner $\mathcal{F}_{learner}$, which performs height classification through $x_{\mathcal{M}_{\theta}}^{rel}$.
Formally, the scale-related loss can be expressed as:
\begin{equation}
\resizebox{0.9\hsize}{!}{$\mathcal{L}_{rel} \left(x_{\mathcal{M}_{\theta}}^{rel}, h; \theta, W_{h} \right)= \; - {{h \cdot {\mathit{\log} {\left( Softmax ( W_{h}^{T} \cdot  x_{\mathcal{M}_{\theta}}^{rel})\right)} }}}$}.
\label{eq2}
\end{equation}
By optimizing $\mathcal{L}_{rel}$, $x_{\mathcal{M}_{\theta}}^{rel}$ can be scale-related through the height level estimation task.

\subsubsection{Feature Disentangling Loss}
Since $x_{\mathcal{M}_{\theta}}^{rel}$ becomes scale-related, we can intuitively make $x_{\mathcal{M}_{\theta}}^{inv}$ scale-invariant by disentangling them.
Specifically, we employ the correlation coefficient analysis~\cite{yang2019survey} to quantify the degree of the disentanglement.
To facilitate the calculation of correlation coefficients, $x_{\mathcal{M}_{\theta}}^{inv}$ and $x_{\mathcal{M}_{\theta}}^{rel}$ are projected as scale-related vector $\hat{x}_{\mathcal{M}_{\theta}}^{rel}$ and scale-invariant vector $\hat{x}_{\mathcal{M}_{\theta}}^{inv}$.
The projection is expressed as:
\begin{equation}
\hat{x}_{\mathcal{M}_{\theta}}^{rel} = W_{rel}^{T} \cdot x_{\mathcal{M}_{\theta}}^{rel}, \hspace{0.15in} \hat{x}_{\mathcal{M}_{\theta}}^{inv} = W_{inv}^{T}\cdot x_{\mathcal{M}_{\theta}}^{inv}.
\label{eq3}
\end{equation}

Then, we calculate the correlation coefficient $\rho$ between $\hat{x}_{\mathcal{M}_{\theta}}^{rel}$ and $\hat{x}_{\mathcal{M}_{\theta}}^{inv}$. The formula is as follows:
\begin{equation}
\rho\left( {\hat{x}_{\mathcal{M}_{\theta}}^{rel},\hat{x}_{\mathcal{M}_{\theta}}^{inv}} \right) = \frac{cov\left(\hat{x}_{\mathcal{M}_{\theta}}^{rel}, \hat{x}_{\mathcal{M}_{\theta}}^{inv} \right)}
{\sqrt{D\left(\hat{x}_{\mathcal{M}_{\theta}}^{rel}\right)} \sqrt{D\left(\hat{x}_{\mathcal{M}_{\theta}}^{inv}\right)}},
\label{eq4}
\end{equation}
where $\mathit{cov}(X,Y)$ is the covariance to measure the correlation between $X$ and $Y$, $D\lbrack X\rbrack$ represents the variance. Since the covariance between two independent random variables should be close to 0, $\rho$ can be utilized as a correlation loss, which can reduce the correlation between scale-related and scale-invariant features.
To facilitate calculation, $\rho^{2}$ is taken as the feature disentangling loss, which can be described as:
\begin{equation}
\begin{aligned}
\mathcal{L}_{\rho} \left(x_{\mathcal{M}_{\theta}}^{rel}, x_{\mathcal{M}_{\theta}}^{inv}; \theta, W_{\mathit{rel}}, W_{\mathit{inv}}\right) 
= \rho^{2}\left( \hat{x}_{\mathcal{M}_{\theta}}^{rel},\hat{x}_{\mathcal{M}_{\theta}}^{inv} \right).
\end{aligned}
\label{eq5}
\end{equation}

\subsection{Adversarial Feature Learning} \label{m3}

Due to the relatively small parameter size of the SIFD, it is prone to insufficient training, resulting in deficient disentangling. Furthermore, the existence of additional $W_{inv}$ and $W_{rel}$ may also cause a reduction in $\rho^{2}$ in their training process. We need to ensure that the decrease in $\rho^{2}$ is attributed to the feature disentanglement rather than to projections $W_{inv}$ and $W_{rel}$.

Inspired by Age-Invariant Adversarial Feature~\cite{liu2022age} for kinship verification,
we employ an Adversarial Feature Learning method to alleviate the above issues for UAV-OD.
Specifically, we freeze detector $\mathcal{M}_{\theta}$ and $\mathcal{F}_{learner}$, then perform gradient reversal~\cite{ganin2016domain}, train $W_{inv}$ and $W_{rel}$ to maximize $\mathcal{L}_{\rho}$. When $\mathcal{L}_{\rho}$ reaches its maximum, we unfreeze $\mathcal{M}_{\theta}$ as well as $\mathcal{F}_{learner}$, and freeze $W_{inv}$ and $W_{rel}$ to minimize $\mathcal{L}_{\rho}$.
Two operations are alternated until $\mathcal{L}_{\rho}$ converges to a minimum.

This training process is an adversarial competition, where one side aims to maximize $\mathcal{L}_{\rho}$, while the other side seeks to minimize it. The entire training process involves iteratively minimizing the maximum feature disentangling loss $\mathcal{L}_{\rho}$. 
To be specific, for every 80 iterations, we take maximum in the first 30 steps and then minimize $\mathcal{L}_{\rho}$ in the next 50 ones. The minimization process is assigned more iterations because our purpose is to accomplish feature disentanglement.
The training scheme can be formulated as follows:
\begin{equation}
\begin{aligned}
\mathcal{L}_{dis} \left(x_{\mathcal{M}_{\theta}}^{rel}, x_{\mathcal{M}_{\theta}}^{inv}; \theta, W_{\mathit{rel}}, W_{\mathit{inv}} \right)= {\min\limits_{\theta}{{~\max\limits_{W_{\mathit{rel}},W_{\mathit{inv}}}} \mathcal{L}_{\rho} }}.
\end{aligned}
\label{eq6}
\end{equation}    


\begin{figure*}[t]
    \centering
    \includegraphics[width=1\linewidth]{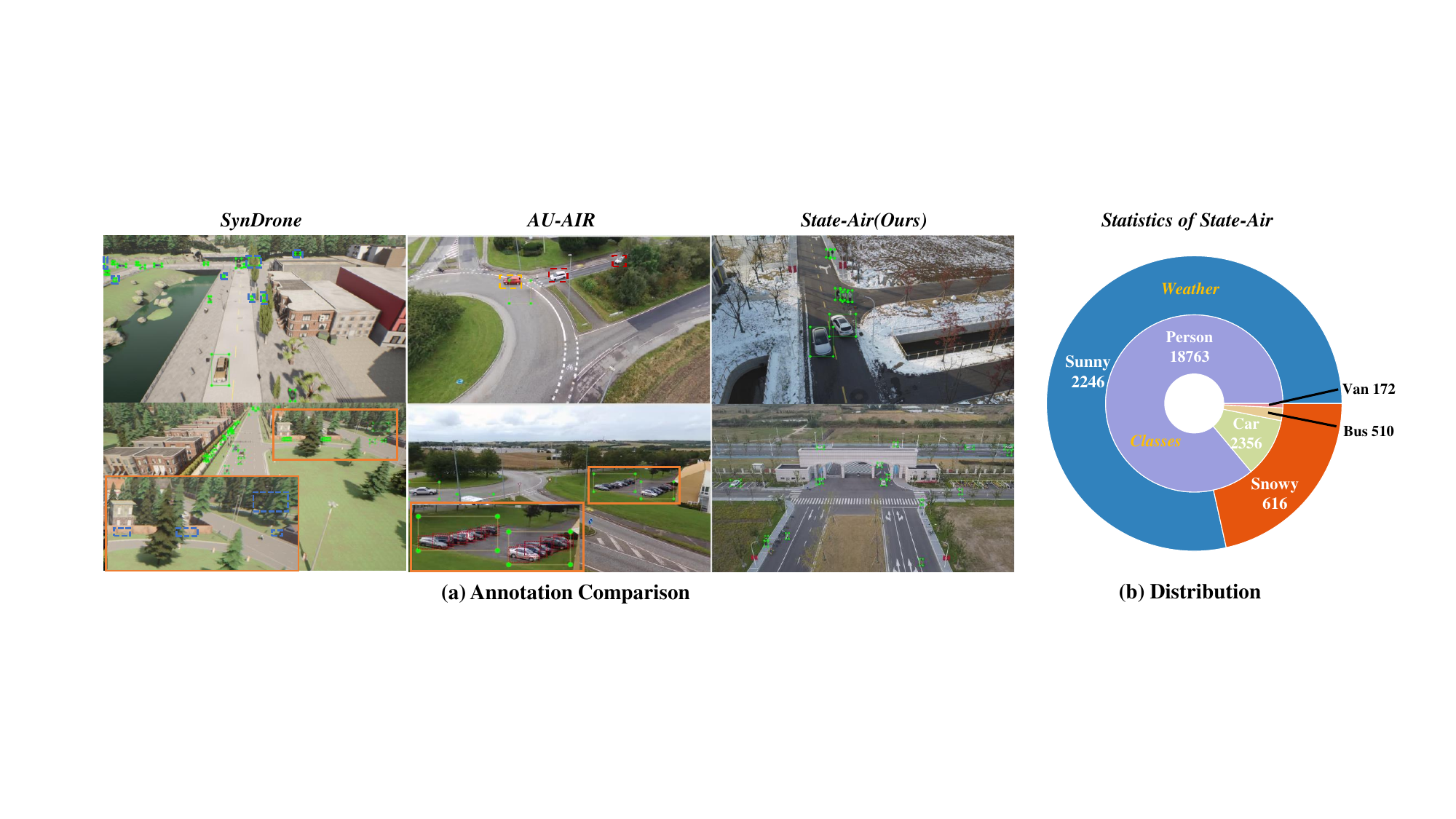}
    \caption{(a) Annotation comparison among State-Air, AU-AIR, and SynDrone. Green: labels given by the datasets; Yellow: revision of incorrect labels; Red: missed labels; Blue: negative labels. (b) State-Air's distribution of scene (outer) and object category (inner).
    }
    \label{fig4}
\end{figure*}

Finally, by integrating scale-related and disentangling loss functions, our overall training target function becomes:
\begin{equation}
\begin{aligned}
\mathcal{L} = \mathcal{L}_{det} + \lambda_{1}\mathcal{L}_{rel}+{\lambda_{2}}\mathcal{L}_{dis},
\end{aligned}
\label{eq8}
\end{equation}
where $\mathcal{L}_{det}$ represents the object detection loss, $\mathcal{L}_{rel}$ and $\mathcal{L}_{dis}$ denote the scale-related loss and the feature disentangling loss, respectively. $\lambda_{1}$ and $\lambda_{2}$ are balancing parameters. Note that $\mathcal{L}_{det}$ can be any object detection loss, making our SIFDAL a plug-and-play approach.

To detect small objects, we utilize the disentangled scale-invariant features $x_{\mathcal{M}_{\theta}}^{inv}$ as input to the high-resolution detection head, as shown in Fig.~\ref{fig2}.
Regarding scale-related features ${x}_{\mathcal{M}_{\theta}}^{rel}$, we opt to discard them as they have the potential to be misleading and detrimental to the detection process. Furthermore, discarding redundant features can reduce computational complexity and slightly improve detection efficiency.

\section{State-Air Dataset} \label{Data}

Existing UAV-OD datasets typically exclude additional data modalities related to the flight, such as those captured by internal sensors.
Nevertheless, flight data (\eg altitude) has the potential to be valuable and could contribute to UAV-OD research.
To this end, we propose an aerial dataset (State-Air) with multi-modal sensor data collected in real-world outdoor environments. The multi-modal information incorporates aerial images and UAV flight status from IMU (height, roll, pitch, and yaw)~\cite{9673756}.
\begin{table}[t]
    \centering
    \caption{Comparison of different UAV-OD datasets. `S/R' represents Synthetic or Real-world data. `MM' denotes multi-modal. `weather' refers to different weather conditions.}
    \renewcommand\arraystretch{1}
    \resizebox{0.5\textwidth}{!}{
    \begin{tabular}{lcccc}
    \toprule
    Dataset   & S/R & Weather & MM & Hegiht[m]\\
    \midrule
    Stanford Drone    & R   & \ding{53}    & \ding{53}  & -  \\
    VisDrone    & R   & \ding{53}          & \ding{53}  & -  \\
    UAV-DT    & R   & \textbf{\checkmark}  & \ding{53}  & -  \\
    \midrule
    AU-AIR    & R   & \ding{53}    & \textbf{\checkmark}  & 5-30  \\
    SynDrone  & S   & \ding{53}    & \textbf{\checkmark} & 20, 50, 80 \\
    \cellcolor[HTML]{DAE8FC}\textbf{State-Air (ours)} & \cellcolor[HTML]{DAE8FC}R   & \cellcolor[HTML]{DAE8FC}\textbf{\checkmark}          & \cellcolor[HTML]{DAE8FC}\textbf{\checkmark}   & \cellcolor[HTML]{DAE8FC}\textbf{5-75} \\
    \bottomrule
    \end{tabular}
    }
    \label{tab:stateair}
\end{table}

The State-Air dataset was collected using DJI Mini2, a micro-UAV~\cite{stankovic2021uav}. We designed an Android APP to obtain and store real-time images containing UAV flight status information through the DJI Mobile SDK. 
Finally, we gathered 2864 aerial images, including 2246 images of sunny days and 616 instances of snowy ones. Each image has a size of 1280 * 720 pixels and contains objects covering a wide variety of scales and shapes.

These aerial images were then annotated for four common object categories: person, car, bus, and van. 
Specifically, objects were initially annotated by Grounding DINO~\cite{liu2023grounding} and subsequently manually proofread. This annotation method is expected to be more accurate and convenient compared to the traditional crowd-sourced image annotations.

Fig.~\ref{fig4} presents sample annotation results for comparing our proposed State-Air with existing two multi-modal UAV datasets, AU-AIR~\cite{bozcan2020air} and SynDrone~\cite{rizzoli2023syndrone}.
It can be observed from Fig.~\ref{fig4} that AU-AIR's annotations tend to be coarse and incorrect, \eg, missing labeling or annotating multiple cars with one box.
Regarding SynDrone, several negative objects blocked by walls or trees are incorrectly labeled.
Furthermore, it is a synthetic imagery dataset, which weakens its applicability to complicated natural scenes. Table~\ref{tab:stateair} presents a detailed statistical comparison of UAV-OD datasets.


\section{Experiments} \label{sec:experiment}
\begin{table*}
\renewcommand\arraystretch{1.5}
\centering
\caption{Comparison of AP and FLOPs on two benchmark datasets by utilizing our approach with various base detectors. Our approach effectively improves detection accuracy and mildly reduces test inference costs on multiple detectors. }
\Large \resizebox{\textwidth}{!}{
\begin{tabular}{cc|cccccc|cccccc|cccccc|c}
\toprule
\multicolumn{2}{c|}{\multirow{2}{*}{Method}} & \multicolumn{6}{c|}{AU-AIR}                    & \multicolumn{6}{c|}{State-Air}       &\multicolumn{6}{c|}{SynDrone}         & \multirow{2}{*}{FLOPs$\downarrow$} \\
\cline{3-20}
\multicolumn{2}{c|}{}                        & $mAP$$\uparrow$   & $AP_{50}$$\uparrow$   & $AP_{75}$$\uparrow$ & $AP_S$$\uparrow$ & $AP_M$$\uparrow$ & $AP_L$$\uparrow$  & 
$mAP$$\uparrow$   & $AP_{50}$$\uparrow$   & $AP_{75}$$\uparrow$ & $AP_S$$\uparrow$ & $AP_M$$\uparrow$ & $AP_L$$\uparrow$  & $mAP$$\uparrow$   & $AP_{50}$$\uparrow$   & $AP_{75}$$\uparrow$ & $AP_S$$\uparrow$ & $AP_M$$\uparrow$ & $AP_L$$\uparrow$  \\
\hline
\multicolumn{1}{c|}{\multirow{3}{*}{RetinaNet}}      & \multicolumn{1}{c|}{baseline}   & 10.7\% & 25.4\% & 5.9\% & - & 6.8\% & 15.9\% 
& 13.8\%  & 30.5\% & 10.6\% & 1.5\%  & 20.0\% & \textbf{48.2\%} & 12.9\% & 22.0\% & 15.6\% & 2.4\%  & 9.2\%  & 36.4\% & 191.423G   \\
\multicolumn{1}{c|}{} 
                                & \multicolumn{1}{c|}{\textbf{SIFDAL}}      & \textbf{12.8\%} & \textbf{27.9\%} & \textbf{6.7\%} & - &  \textbf{8.4\%} &  \textbf{16.2\%} &  \textbf{15.0\%} &  \textbf{31.9\%} &  \textbf{12.0\%} &  \textbf{4.7\%}  &  \textbf{21.4\%} &  47.8\% &    \textbf{15.4\%} &  \textbf{24.1\%} &  \textbf{19.4\%} &  \textbf{5.9\%}  &  \textbf{11.7\%} &  \textbf{46.7\%} & 185.217G                      \\
                                \multicolumn{1}{c|}{} 
                                & \multicolumn{1}{c|}{\cellcolor[HTML]{DAE8FC}\textbf{$\pm \Delta$}}      & \cellcolor[HTML]{DAE8FC}\textbf{2.1\%} & \cellcolor[HTML]{DAE8FC}\textbf{2.5\%} & \cellcolor[HTML]{DAE8FC}\textbf{0.8\%} & \cellcolor[HTML]{DAE8FC}- & \cellcolor[HTML]{DAE8FC} \textbf{1.6\%} & \cellcolor[HTML]{DAE8FC} \textbf{0.3\%} & \cellcolor[HTML]{DAE8FC} \textbf{1.2\%} & \cellcolor[HTML]{DAE8FC} \textbf{1.4\%} & \cellcolor[HTML]{DAE8FC} \textbf{1.4\%} & \cellcolor[HTML]{DAE8FC} \textbf{3.2\%}  & \cellcolor[HTML]{DAE8FC} \textbf{1.4\%} & \cellcolor[HTML]{DAE8FC} -0.4\%  &    \cellcolor[HTML]{DAE8FC}\textbf{2.5\%} & \cellcolor[HTML]{DAE8FC} \textbf{2.1\%} & \cellcolor[HTML]{DAE8FC} \textbf{3.8\%} & \cellcolor[HTML]{DAE8FC} \textbf{3.5\%}  & \cellcolor[HTML]{DAE8FC} \textbf{2.5\%} & \cellcolor[HTML]{DAE8FC} \textbf{10.3\%}&  -                     \\
\hline
\multicolumn{1}{c|}{\multirow{3}{*}{EfficientDet}}   & baseline   & 11.0\% & 29.1\% & 5.4\% & - & 6.6\% & 16.0\% 
& 27.0\%  & 63.2\% & 25.9\% & 7.6\%  & 40.9\% & 42.1\% & 20.8\% & 31.9\% & 22.9\% & 2.0\%  & 29.2\% & 40.6\% & 22.394G                   \\
                                \multicolumn{1}{c|}{} 
                                &  \textbf{SIFDAL}      &  \textbf{13.1\%} &  \textbf{30.1\%} &  \textbf{7.2\%} & - &  \textbf{8.7\%} &  \textbf{17.5\%} &  \textbf{31.9\%}  &  \textbf{66.3\%} &  \textbf{26.5\%} &  \textbf{10.8\%} &  \textbf{47.2\%} &  \textbf{55.5\%} &  \textbf{21.7\%} &  \textbf{32.6\%} &  \textbf{23.7\%} &  \textbf{6.2\%}  &  \textbf{30.6\%} &  \textbf{43.2\%}& 21.756G                       \\
                                \multicolumn{1}{c|}{} 
                                & \multicolumn{1}{c|}{\cellcolor[HTML]{DAE8FC}\textbf{$\pm \Delta$}}      & \cellcolor[HTML]{DAE8FC}\textbf{2.1\%} & \cellcolor[HTML]{DAE8FC}\textbf{1.0\%} & \cellcolor[HTML]{DAE8FC}\textbf{1.8\%} & \cellcolor[HTML]{DAE8FC}- & \cellcolor[HTML]{DAE8FC} \textbf{2.1\%} & \cellcolor[HTML]{DAE8FC} \textbf{1.5\%} & \cellcolor[HTML]{DAE8FC} \textbf{4.9\%} & \cellcolor[HTML]{DAE8FC} \textbf{3.1\%} & \cellcolor[HTML]{DAE8FC} \textbf{0.6\%} & \cellcolor[HTML]{DAE8FC} \textbf{3.2\%}  & \cellcolor[HTML]{DAE8FC} \textbf{6.3\%} & \cellcolor[HTML]{DAE8FC} \textbf{13.4\%} & \cellcolor[HTML]{DAE8FC} \textbf{0.9\%} & \cellcolor[HTML]{DAE8FC} \textbf{0.7\%} & \cellcolor[HTML]{DAE8FC} \textbf{0.8\%} & \cellcolor[HTML]{DAE8FC} \textbf{4.2\%} & \cellcolor[HTML]{DAE8FC} \textbf{1.4\%} & \cellcolor[HTML]{DAE8FC} \textbf{2.6\%}&  -                     \\
\hline                                
\multicolumn{1}{c|}{\multirow{3}{*}{YOLOv7}}         & baseline   & 13.3\% & 34.6\% & 7.4\% & - & 9.4\% & 16.4\%  
& 31.5\%  & 81.0\% & 16.5\% & 23.9\% & 38.1\% & 40.2\% & 31.8\% & 65.2\% & 34.3\% & 13.8\% & 39.3\% & 41.0\% & 106.472G    \\
                                \multicolumn{1}{c|}{} 
                                &  \textbf{SIFDAL}      &  \textbf{14.1\%} &  \textbf{36.9\%} &  \textbf{8.4\%} & - & 9.4\% &  \textbf{17.5\%} &  \textbf{37.5\%} &  \textbf{90.0\%} &  \textbf{37.2\%} &  \textbf{28.4\%} &  \textbf{43.8\%} &  \textbf{67.6\%} &  \textbf{34.6\%} &  \textbf{68.3\%} &  \textbf{40.2\%} &  \textbf{18.4\%} &  \textbf{42.2\%} &  \textbf{46.7\%} &       103.136G          \\
                                \multicolumn{1}{c|}{} 
                                & \multicolumn{1}{c|}{\cellcolor[HTML]{DAE8FC}\textbf{$\pm \Delta$}}      & \cellcolor[HTML]{DAE8FC}\textbf{0.8\%} & \cellcolor[HTML]{DAE8FC}\textbf{2.3\%} & \cellcolor[HTML]{DAE8FC}\textbf{1.0\%} & \cellcolor[HTML]{DAE8FC}- & \cellcolor[HTML]{DAE8FC} \textbf{0.0\%} & \cellcolor[HTML]{DAE8FC} \textbf{1.1\%} & \cellcolor[HTML]{DAE8FC} \textbf{6.0\%} & \cellcolor[HTML]{DAE8FC} \textbf{9.0\%} & \cellcolor[HTML]{DAE8FC} \textbf{20.7\%} & \cellcolor[HTML]{DAE8FC} \textbf{4.5\%}  & \cellcolor[HTML]{DAE8FC} \textbf{5.7\%} & \cellcolor[HTML]{DAE8FC} \textbf{27.4\%} & \cellcolor[HTML]{DAE8FC} \textbf{2.8\%} & \cellcolor[HTML]{DAE8FC} \textbf{3.1\%} & \cellcolor[HTML]{DAE8FC} \textbf{5.9\%} & \cellcolor[HTML]{DAE8FC} \textbf{4.6\%} & \cellcolor[HTML]{DAE8FC} \textbf{2.9\%} & \cellcolor[HTML]{DAE8FC} \textbf{5.7\%}&  -                     \\
\bottomrule
\end{tabular}
}
\label{tab:comparison}
\end{table*}

\begin{table}[t]
\centering
\renewcommand\arraystretch{1.3}
\caption{Comparison with other state-of-the-art (SoTA) methods. `Ours' denotes YOLOv7 utilized our SIFDAL.}
\resizebox{\linewidth}{!}{
\begin{tabular}{c|cc|ccc}
\toprule
Dataset                    & Method     & Publication & $mAP$   & $AP_{50}$   & $AP_{75}$  \\
\hline
\multirow{4}{*}{AU-AIR}    & CEASC~\cite{CEASC}      & CVPR2023    & 13.2\% & 32.6\%  & 7.5\%  \\
                       & CZDetector~\cite{CZDetector} & CVPR2023    & 11.8\% & 29.3\%  & 5.3\%  \\
                           & RT-DETR~\cite{lv2023detrs}    & CVPR2024    & 12.5\% & 29.7\% & \textbf{9.0\%}  \\
                           & D-FINE~\cite{peng2024d}    & arxiv2024    & 12.7\% & 30.0\%  & 6.9\% \\
                           & \cellcolor[HTML]{DAE8FC}\textbf{Ours}       & \cellcolor[HTML]{DAE8FC}-           & \cellcolor[HTML]{DAE8FC}\textbf{14.1\%} & \cellcolor[HTML]{DAE8FC}\textbf{36.9\%}  & \cellcolor[HTML]{DAE8FC}8.4\%  \\
\hline                           
\multirow{4}{*}{State-Air} & CEASC~\cite{CEASC}      & CVPR2023    & 36.0\% & 87.8\%  & 27.5\% \\
                           & CZDetector~\cite{CZDetector} & CVPR2023    & 29.7\% & 67.5\%  & 27.2\% \\
                           & RT-DETR~\cite{lv2023detrs}    & CVPR2024    & 35.9\% & 89.9\%  & 31.6\% \\
                           & D-FINE~\cite{peng2024d}    & arxiv2024    & 33.5\% & 83.6\%  & 34.9\% \\
                           & \cellcolor[HTML]{DAE8FC}\textbf{Ours}       & \cellcolor[HTML]{DAE8FC}-           & \cellcolor[HTML]{DAE8FC}\textbf{37.5\%} & \cellcolor[HTML]{DAE8FC}\textbf{90.0\%}  & \cellcolor[HTML]{DAE8FC}\textbf{37.2\%} \\
\bottomrule
\end{tabular}
}
\label{tab:sota}
\end{table}

\subsection{Datasets and Evaluation Metrics}
We adopted three datasets that have UAV altitude information for experiments (AU-AIR~\cite{bozcan2020air}, SynDrone~\cite{rizzoli2023syndrone}, and our proposed State-Air). 

AU-AIR~\cite{bozcan2020air} collected 32,823 video frames and manually annotated 8 object categories, totaling 13,203 objects. Additionally, it provides various parameters from the UAV, including time, GPS, altitude, and linear velocity. This dataset features both IMU parameters and RGB images, making it a multi-modal dataset for UAV-based object detection.

SynDrone~\cite{rizzoli2023syndrone} is built on the CARLA~\cite{dosovitskiy2017carla} simulator. Utilizing the high-quality simulation technology of Unreal Engine 4 (UE4), it collected a total of 60,000 pairs of multimodal images across 28 categories, including RGB, depth, and radar. It supports both object detection and instance segmentation tasks, making it one of the largest-scale multi-modal UAV-based object detection datasets available.

Following the same evaluation metrics of previous work~\cite{zou2023object,10440361,10737701}, we employed the Average Precision (AP)~\cite{everingham2010pascal} as the evaluation metrics on accuracy. Specifically, mAP is averaged on ten intersection over union (IoU) values of [0.50 : 0.05 : 0.95], $AP_{50}$ and $AP_{75}$ are computed at the single IoU of 0.5 and 0.75, respectively, $AP_{S}$, $AP_{M}$ , and $AP_{L}$ are computed for targets of different sizes, where the pixel sizes of the small targets are less than 32 × 32, the pixel sizes of the medium targets are in the range of 32 × 32–96 × 96, and the pixel sizes of the large targets are larger than 96 × 96. We also utilize FLOPs to measure the efficiency of our proposed method.

\subsection{Experimental Setup} \label{experiment}
\subsubsection{Baselines} We adopted YOLOv7-L~\cite{wang2023yolov7}, EfficientDet-d2~\cite{tan2020efficientdet} and RetinaNet~\cite{lin2017focal} with ResNet50~\cite{he2016deep} as the baseline models. They are representative FPN-based detection frameworks and we tested our proposed SIFDAL on them.
It's worth mentioning that in the \textbf{VisDrone Challenge of ICCV2023} (the authoritative challenge in UAV-OD), the champion team utilized YOLOv7 as their framework. Therefore, the experiments in this paper primarily focus on it.

\subsubsection{Training Settings} All experiments were conducted in Pytorch with 3 $\times$ NVIDIA RTX 3090 GPUs. We trained the framework for 200 epochs with a batch size of 128. All detectors were trained using an Adam optimizer~\cite{kingma2014adam} with a momentum of 0.937, and the learning rate was initialized as 0.001 with a cosine decay~\cite{loshchilovstochastic}. Modules in SIFD were optimized using AdamW~\cite{loshchilov2017decoupled} with the same setting as Adam. Both $\lambda_{1}$ and $\lambda_{2}$ were set to 0.3. The height category number of State-Air and AU-AIR were determined by Silhouette Coefficient~\cite{rousseeuw1987silhouettes,pelleg2000x} during clustering. Finally, the number of height categories for the two datasets were 5 and 8, respectively. We fix random seed to 18 to ensure the experiment's reproducibility.


\subsection{Experimental Results and Analyses}

\subsubsection{Effectiveness of SIFDAL}
As demonstrated in Table~\ref{tab:comparison}, applying SIFDAL to FPN-based frameworks leads to an improvement of more than 1.0\% in the $mAP$ scores in the majority of cases.

\begin{table}[t]
    \centering
    \caption{Impact of utilizing scale-related features and scale-invariant features with YOLOv7 as the base detector.}
    \renewcommand\arraystretch{1}
    \resizebox{\linewidth}{!}{
    \begin{tabular}{c>{\centering\arraybackslash}p{0.7cm}>{\centering\arraybackslash}p{0.7cm}ccc}
        \toprule  
        Datasets       & \centering $x_{\mathcal{M}_{\theta}}^{inv}$ & \centering $x_{\mathcal{M}_{\theta}}^{rel}$ & $mAP$ & $AP_{50}$   & $AP_{75}$   \\
        \hline
        \multirow{4}{*}{AU-AIR}  &   &   & 13.3\%  & 34.6\% & 7.4\% \\
                                    & \centering \cellcolor[HTML]{DAE8FC}\checkmark                         &             \cellcolor[HTML]{DAE8FC}         & \cellcolor[HTML]{DAE8FC}\textbf{14.1\%}    & \cellcolor[HTML]{DAE8FC}\textbf{36.9\%} & \cellcolor[HTML]{DAE8FC}\textbf{8.4\%} \\
                                    &                & \centering \checkmark    & 12.4\%         & 31.8\% & 6.9\% \\
                                    & \centering \checkmark                         & \centering \checkmark   & 13.8\%          & 35.4\% & 7.1\% \\
        \hline
        \multirow{4}{*}{State-Air}  &   &   & 31.5\%   & 81.0\% & 16.5\% \\
                                    & \centering \cellcolor[HTML]{DAE8FC}\checkmark                         &             \cellcolor[HTML]{DAE8FC}         & \cellcolor[HTML]{DAE8FC}\textbf{37.5\%}    & \cellcolor[HTML]{DAE8FC}\textbf{90.0\%} & \cellcolor[HTML]{DAE8FC}\textbf{37.2\%} \\
                                    &                & \centering \checkmark    & 28.7\%           & 75.1\% & 14.7\% \\
                                    & \centering \checkmark                         & \centering \checkmark     & 35.4\%         & 83.6\% & 29.1\% \\
        \bottomrule
    \end{tabular}
    }
    \label{tab:ablation2}
\end{table}

\textbf{Results on AU-AIR.} SIFDAL increased the $AP_{50}$ of YOLOv7, EfficientDet, and RetinaNet by 2.3\%, 1.0\%, and 2.5\%, respectively. 
Due to the imprecise labeling of small objects in the AU-AIR dataset, the $AP_S$ test results are deemed unreliable. Therefore, these results are not recorded.
It can be noted that the accuracies of most baseline detectors are relatively low and the differences between them are marginal. The reason may be that the annotations of AU-AIR are not precise enough. In other words, the results on AU-AIR may not accurately reflect the model's true performance.

\textbf{Results on State-Air.} SIFDAL exhibits more significant improvements on State-Air than on the other two datasets.
Specifically, it achieves remarkable $AP_{50}$ and $AP_{75}$ gains of 9.0\% and 20.7\% on YOLOv7, respectively. In particular, for small objects, our SIFDAL leds to an increase in the $AP_S$ metric for the three detectors by 4.5\%, 3.2\%, and 3.2\%, respectively.
The remarkable improvements reveal the effectiveness of our approach. Moreover, all the detectors achieve their best performance on State-Air, which may be attributed to better precise and accurate labels than AU-AIR.

\textbf{Results on SynDrone.}We also adopted SynDrone~\cite{rizzoli2023syndrone} for experiments to validate our SIFDAL. SynDrone is a multi-modal synthetic benchmark dataset containing both images and 3D data taken at multiple flying heights. It includes 28 classes of pixel-level labeling and object-level annotations for semantic segmentation and object detection. The height category number of SynDrone was fixed at 3 (20m, 50m, 80m). Our SIFDAL can achieve remarkable performance with $AP_{50}$ improvements by 3.1\%, 0.7\% and 2.1\% on three single-stage detectors, respectively. 
Furthermore, it could be found that most baseline approaches attain moderate scores on SynDrone among the three benchmarks. The reason may be that the existence of negative samples in SynDrone misguides the training and test processes. Furthermore, as a synthetic dataset, SynDrone lacks authenticity to some extent and its practical application tends to be limited.

\subsubsection{Comparison with SoTA Methods}
As demonstrated in Table~\ref{tab:sota}, we selected the CEASC~\cite{CEASC}, RT-DETR~\cite{lv2023detrs}, D-FINE~\cite{peng2024d}, and CZdetector\cite{CZDetector} for comparison, both of which are of similar scale to YOLOv7. Compared to other SoTA models, our YOLOv7 with SIFDAL achieves a higher $mAP$ than RT-DETR by 1.6\% on AU-AIR. It also surpasses RT-DETR and CEASC by 1.5\% and 1.6\% in $mAP$ on State-Air, respectively. 

\subsection{Ablation Studies}

In this section, we examined the impact of scale-related and scale-invariant features, validated the effectiveness of each module, and assessed the influence of SIFDAL at different layers on AU-AIR and State-Air datasets with YOLOv7 as the base detector. We also validated the optimal balancing factors and tested the impact of applying SIFD across different FPN layers on the final results. Additionally, we explored other disentangle methods and the effect of adversarial learning on training stability.

\begin{table}[t]
    \centering
    \caption{Effectiveness analysis on our proposed SIFD and AFL with YOLOv7 as the base detector. }
    \renewcommand\arraystretch{1.2}
    \resizebox{\linewidth}{!}{
    \begin{tabular}{c>{\centering\arraybackslash}p{0.7cm}>{\centering\arraybackslash}p{0.7cm}ccc}
    \toprule
    Datasets                   & \centering SIFD & \centering AFL 
    & $mAP$ & $AP_{50}$   & $AP_{75}$  \\
    \hline
    \multirow{3}{*}{AU-AIR}    &      &     &13.3\%   & 34.6\% & 7.4\%  \\
                               & \centering \checkmark    &        &13.4\%  & 35.7\% & 7.9\%  \\
                               & \centering \cellcolor[HTML]{DAE8FC}\checkmark    & \centering \cellcolor[HTML]{DAE8FC}\checkmark   &\cellcolor[HTML]{DAE8FC}\textbf{14.1\%} & \cellcolor[HTML]{DAE8FC}\textbf{36.9\%} & \cellcolor[HTML]{DAE8FC}\textbf{8.4\%}  \\
    \hline
    \multirow{3}{*}{State-Air} &      &     &31.5\%   & 81.0\% & 16.5\% \\
                               & \centering \checkmark    &    &35.9\%    & 87.6\% & 26.3\% \\
                               & \centering \cellcolor[HTML]{DAE8FC}\checkmark    & \centering \cellcolor[HTML]{DAE8FC}\cellcolor[HTML]{DAE8FC}\checkmark   & \cellcolor[HTML]{DAE8FC}\textbf{37.5\%}& \cellcolor[HTML]{DAE8FC}\textbf{90.0\%} & \cellcolor[HTML]{DAE8FC}\textbf{37.2\%} \\
    \bottomrule
    \end{tabular}
    }
    \label{tab:ablation3}
\end{table}

\subsubsection{Impact of Scale-Related and Scale-Invariant Features}
We leveraged the model without feature disentanglement as the baseline and conducted an experiment to verify the optimal features for UAV-OD, as reported in Table~\ref{tab:ablation2}.
The most notable improvement can be achieved by merely utilizing disentangled scale-invariant features $x_{\mathcal{M}_{\theta}}^{inv}$, with an increase of 2.3\% and 9.0\% on AU-AIR and State-Air, respectively.
Conversely, only using scale-related features $x_{\mathcal{M}_{\theta}}^{rel}$ results in the worst performance, even inferior to the baseline.

Applying both two sets of features results in a 2.6\% accuracy gain on State-Air, which is inferior to leveraging $x_{\mathcal{M}_{\theta}}^{inv}$ but outperforms the option to solely utilize $x_{\mathcal{M}_{\theta}}^{rel}$. The reason can be that scale-invariant features contribute to the robust UAV-OD while scale-related features could interfere with the detection process.
The experimental results suggest that scale-invariant features tend to be discriminative and can benefit detection performance.




\begin{table}[t]
    \centering
    \caption{Influence of SIFDAL utilization with YOLOv7 as the base detector. `Location' denotes SIFDAL on different layers of FPN.}
    \renewcommand\arraystretch{1.2}
    \resizebox{\linewidth}{!}{
    \begin{tabular}{c>{\centering\arraybackslash}p{1.6cm}>{\centering\arraybackslash}p{1.1cm}>{\centering\arraybackslash}p{1.1cm}>{\centering\arraybackslash}p{1.1cm}}
        \toprule
         Datasets                & Location & $mAP$  & $AP_{50}$   & $AP_{75}$   \\
        \hline
        \multirow{4}{*}{AU-AIR}  & -          & 13.3\%  & 34.6\% & 7.4\%  \\
                                 & P5        & 13.4\%   & 34.9\% & 7.4\%  \\
                                 & P4       & 13.8\%   & 35.0\% & 8.0\%  \\
                                 & \cellcolor[HTML]{DAE8FC} P3       & \cellcolor[HTML]{DAE8FC}\textbf{14.1\%}        & \cellcolor[HTML]{DAE8FC}\textbf{36.9\%}& \cellcolor[HTML]{DAE8FC}\textbf{8.4\%}  \\
        \hline
        \multirow{4}{*}{State-Air} & -     & 31.5\% & 81.0\% & 16.5\% \\
                       & P5     & 36.2\% & 84.8\% & 25.4\% \\
                       & P4    & 35.6\% & 87.8\% & 24.6\% \\
                       & \cellcolor[HTML]{DAE8FC}P3  & \cellcolor[HTML]{DAE8FC}\textbf{37.5\%}  & \cellcolor[HTML]{DAE8FC}\textbf{90.0\%} & \cellcolor[HTML]{DAE8FC}\textbf{37.2\%}\\
        \bottomrule
    \end{tabular}
    }
    \label{tab:location}
\end{table}

\begin{figure}[t]
\centering
\includegraphics[width=0.5\textwidth]{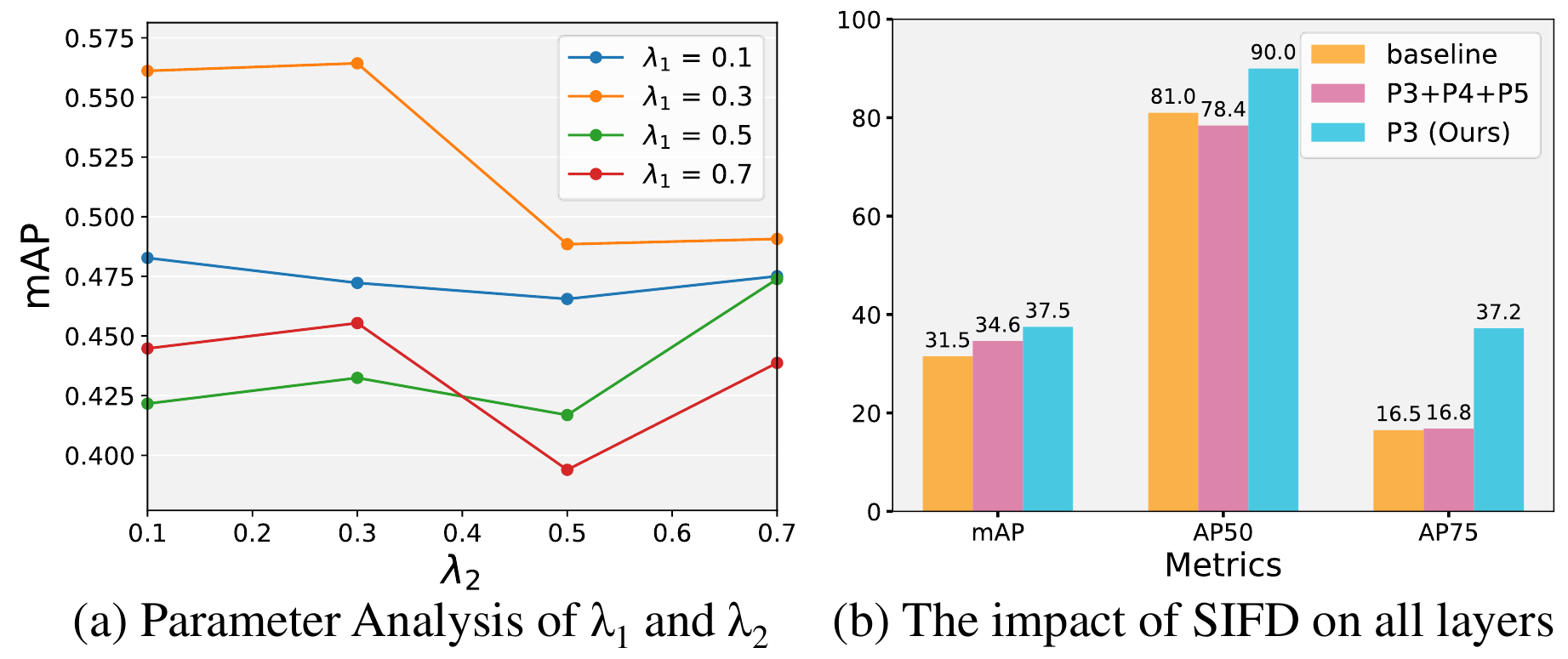}
\caption{(a) Parameter Analysis of $\lambda_1$ and $\lambda_2$. (b) The impact of employing the SIFD module to each layer of the FPN with YOLOv7-L.}
\label{balance}
\end{figure}

\subsubsection{Effectiveness of SIFD and AFL}

We conducted an ablation analysis of SIFD and AFL and assessed their impact on the final results in Table~\ref{tab:ablation3}. 
By employing SIFD, the $AP_{50}$ on AU-AIR and State-Air increase by 1.1\% and 6.6\%, respectively. The performance gain can be attributed to employing scale-invariant features by SIFD.
After adopting AFL, the detection accuracy further grows by 1.2\% and 2.4\% on two datasets, respectively.
It can be concluded that our AFL method can achieve more thorough feature disentanglement.
The experimental results demonstrate that both SIFD and AFL are effective and contribute to the final performance.

\subsubsection{SIFD at Different FPN layers}
We investigated the influence of the SIFDAL location on the effectiveness of disentanglement and conducted experiments at different resolution layers of FPN. 
The resolution of the feature map decreases sequentially from P3 to P5. The experimental results are demonstrated in Table~\ref{tab:location}. 
We can observe that the detection accuracy gradually increases from layer P5 to P3, reaching its peak in the high-resolution layer (P3). For example, the SIFD module applied to the P3, P4, and P5 layers improved the $AP_{50}$ scores by 2.3\%, 0.4\%, and 0.3\% on AU-AIR, respectively. And on the State-Air, the $mAP$ scores for the three layers increased by 6.0\%, 4.1\%, and 4.7\%, respectively.
This result indicates the significance of guiding high-resolution detection heads to leverage the scale-invariant features for UAV-OD tasks.

\subsubsection{Analysis of Balance Factors} \label{Analysis}
To ascertain the optimal balance factor values for model training, we conducted experiments by varying the two balance factors, $\lambda_1$ and $\lambda_2$, within the loss function. The model was trained with different parameters, and the mean Average Precision (mAP) value of the model at the 50th epoch was recorded. The set of parameters yielding the highest accuracy at the 50th epoch was selected as the final training parameters. The experimental results are illustrated in Figure~\ref{balance} (a). 

When the value of $\lambda_1$ is set to 0.1, the model's performance is observed to be subpar. However, setting $\lambda_1$ to an excessively large value may lead to degradation in the model's performance. The reason may be that a too small value (0.1) has little effect on the overall loss function, while a too large value (0.5 or 0.7) can cause the model to focus on feature disentanglement rather than object detection. As for $\lambda_2$, there are obvious fluctuations in the model performance when $\lambda_2$ gets large, which may result from too strong adversarial training.
This phenomenon indicates that a smaller $\lambda_2$ can make the model performance relatively stable.
It is also evident that the model achieves its peak performance when $\lambda_1$ = $\lambda_2$ = 0.3. Therefore, we ultimately adopted this set of values in our method.

\subsubsection{Applying SIFD Module to All Layers of FPN}

We further applied the SIFD module to all layers of the feature pyramid, and the experimental results are shown in Figure~\ref{balance} (b).
The experimental results indicate that when employing the SIFD module in all layers of the FPN, there is a decrease in model accuracy. This could be due to the negative impact of using only scale-invariant features in each layer on multi-scale feature fusion. 
If the multi-scale feature fusion is impaired, the feature discriminability will be weaken and model accuracy will decrease.
Conversely, by only utilizing the SIFD module in the P3 layer, the scale-related features can still participate in multi-scale feature fusion, preserving the original function of the FPN. Therefore, we only applying our SIFD module to the P3 layer in our proposed approach.


\begin{table}[t]
\centering

\caption{Comparison of different disentangling methods on AP and FLOPs with YOLOv7-L.}
\renewcommand\arraystretch{1.2}
\resizebox{\linewidth}{!}{
\begin{tabular}{ccccc}
\toprule
Dataset       & Method &$AP_{0.5}$ & $AP_{0.75}$ & FLOPs \\
\hline
\multirow{3}{*}{AU-AIR}    & -                      & 34.6\%         & 7.4\%  & 106.472G\\
                           & Convlution                   & 35.9\% & 7.3\% & 106.908G \\
                           & \cellcolor[HTML]{DAE8FC}Channel Splitting                  & \cellcolor[HTML]{DAE8FC}\textbf{36.9\%} & \cellcolor[HTML]{DAE8FC}\textbf{8.4\%} & \cellcolor[HTML]{DAE8FC}\textbf{103.136G} \\
\hline                           
\multirow{3}{*}{State-Air} & -                      & 81.0\%         & 16.5\% & 106.472G\\
                           & Convlution                   & 88.9\%         & 27.5\%  & 106.908G\\
                           & \cellcolor[HTML]{DAE8FC}Channel Splitting & \cellcolor[HTML]{DAE8FC}\textbf{90.0\%}         & \cellcolor[HTML]{DAE8FC}\textbf{37.2\%} &\cellcolor[HTML]{DAE8FC}\textbf{103.136G} \\
\bottomrule
\end{tabular}
}

\label{tab:appendix2}
\end{table}

\begin{figure}[t]
    \centering
    \includegraphics[width=1\linewidth]{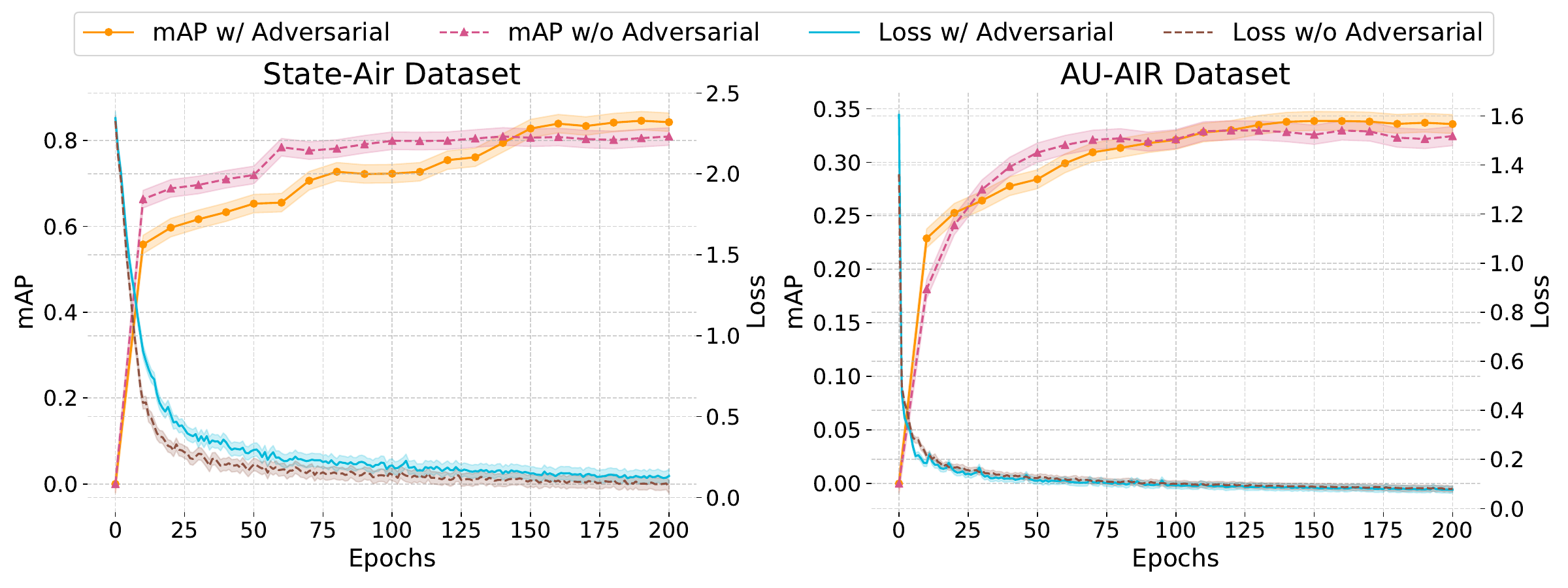}
    \caption{Curves in mAP and loss during training on AU-AIR and State Air datasets.} 
    \label{curves} 
\end{figure}


\subsubsection{Convlutional Disentanglement}

We investigated an alternative feature separation method which replaces channel splitting with two convolutional layers. The experimental results are demonstrated in Table~\ref{tab:appendix2}. It can be observed that the convolutional feature separation method can also improve model accuracy, but the improvement is not as significant as the direct channel spliting.
Additionally, the introduction of two extra convolutional layers also results in computational costs. Therefore, we ultimately chose the channel splitting method, which is simple yet effective.

\subsubsection{Impact of Adversarial Learning}

Since Adversarial Learning may cause training fluctuations, we conducted ablation analysis on the impact of AFL on the training process. The analysis results are shown in Fig.~\ref{curves}. 
It can be observed that regardless of whether the AFL strategy is applied, the mAP curves show a stable increase, while the loss curves continue to steadily decrease. 
Although the convergence speed of the model slightly decreases after utilizing adversarial training, the final accuracy of the model evidently rises. 
This result indicates that our AFL method doesn't have evident impact on the training stability, while it can effectively improve model performance by enhancing the feature disentanglement.

\subsection{Setting scale categories via Bbox size.}
\begin{figure}[t]
\centering
\includegraphics[width=1\linewidth]{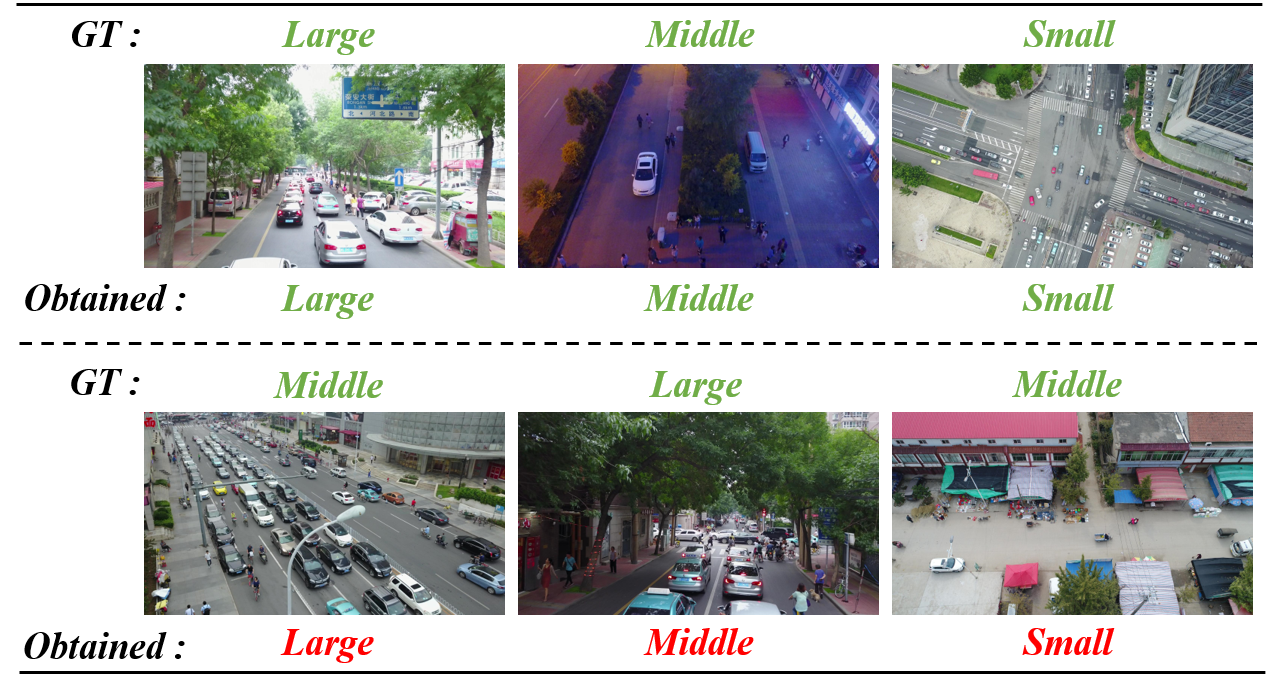}
\caption{The situation of clustering errors based on Bbox size. The `Obtained' indicates the labels obtained from clustering, and the `GT' represents the true labels of images. Due to the diversity of gimbal angles, directly clustering via bounding box (Bbox) size can not yield accurate scale labels.}
\label{fig9}
\end{figure}

The size of bounding boxes (Bbox) within the same category can indirectly reflect the current flight altitude of the UAV and intuitively represent the object scale. We selected car objects and performed clustering via the average size of car Bbox in the current images. The clustering results are shown in Fig~\ref{fig9}. We found that this clustering method resulted in many incorrect labels. For example, the first image in the second row should belong to the medium-scale object category, but it was clustered as a large-scale object. The reason is that the UAV's gimbal angle varies significantly, and even at a low flight altitude, there can be many small objects at a considerable horizontal distance. Therefore, we selected reliable flight altitude annotations as the basis for clustering.



\begin{figure*}[t]
    \centering
    \includegraphics[width=1\linewidth]{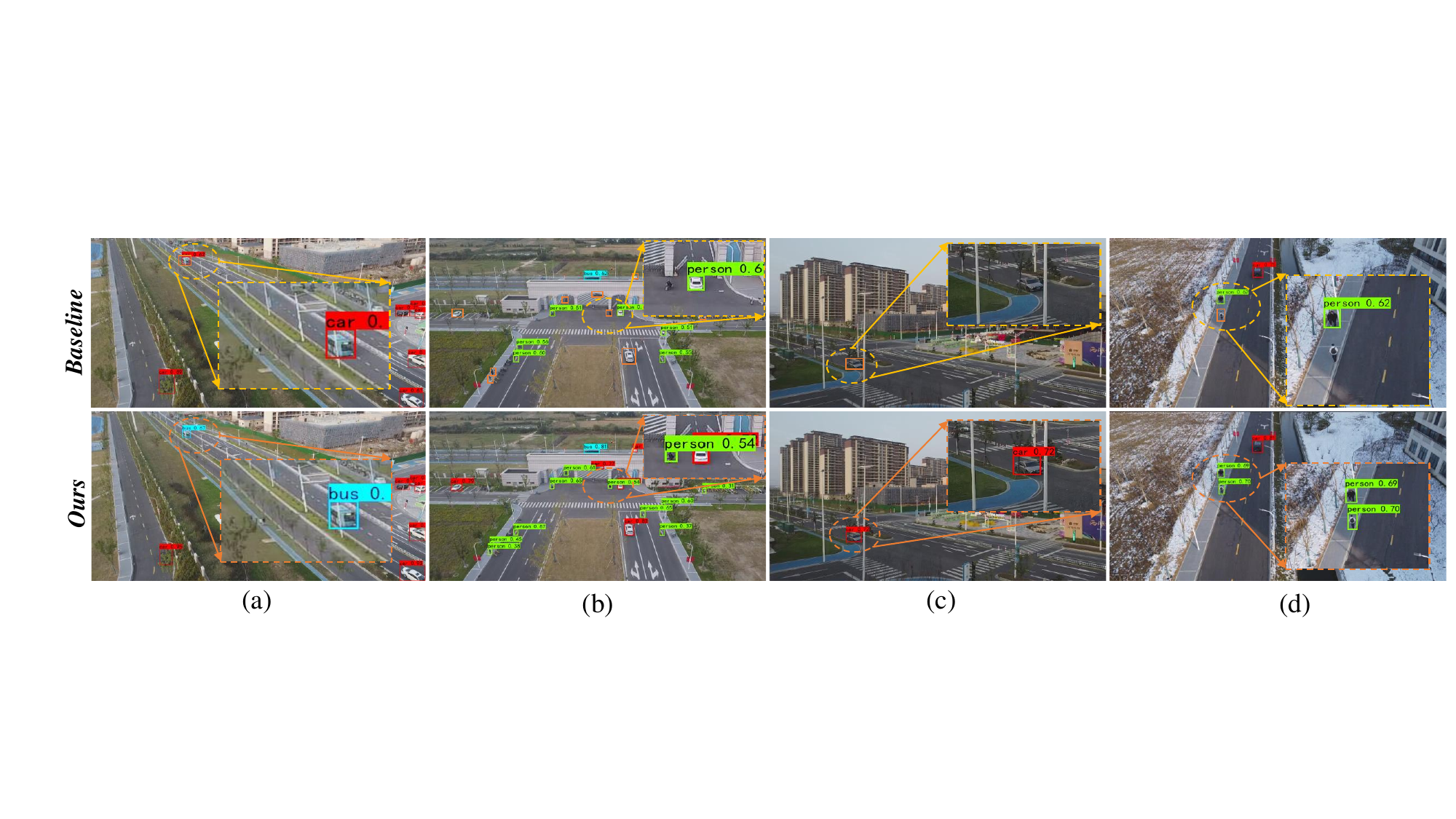}
    \caption{
    Visualization of detection results of SIFDAL and baselines on State-Air. 
    Red, green, and blue boxes represent car, person, and bus predictions, respectively. Orange boxes denote false or missing detection results.
    }
    \label{fig6}
\end{figure*}

\begin{figure*}[t]
    \centering
    \includegraphics[width=1\linewidth]{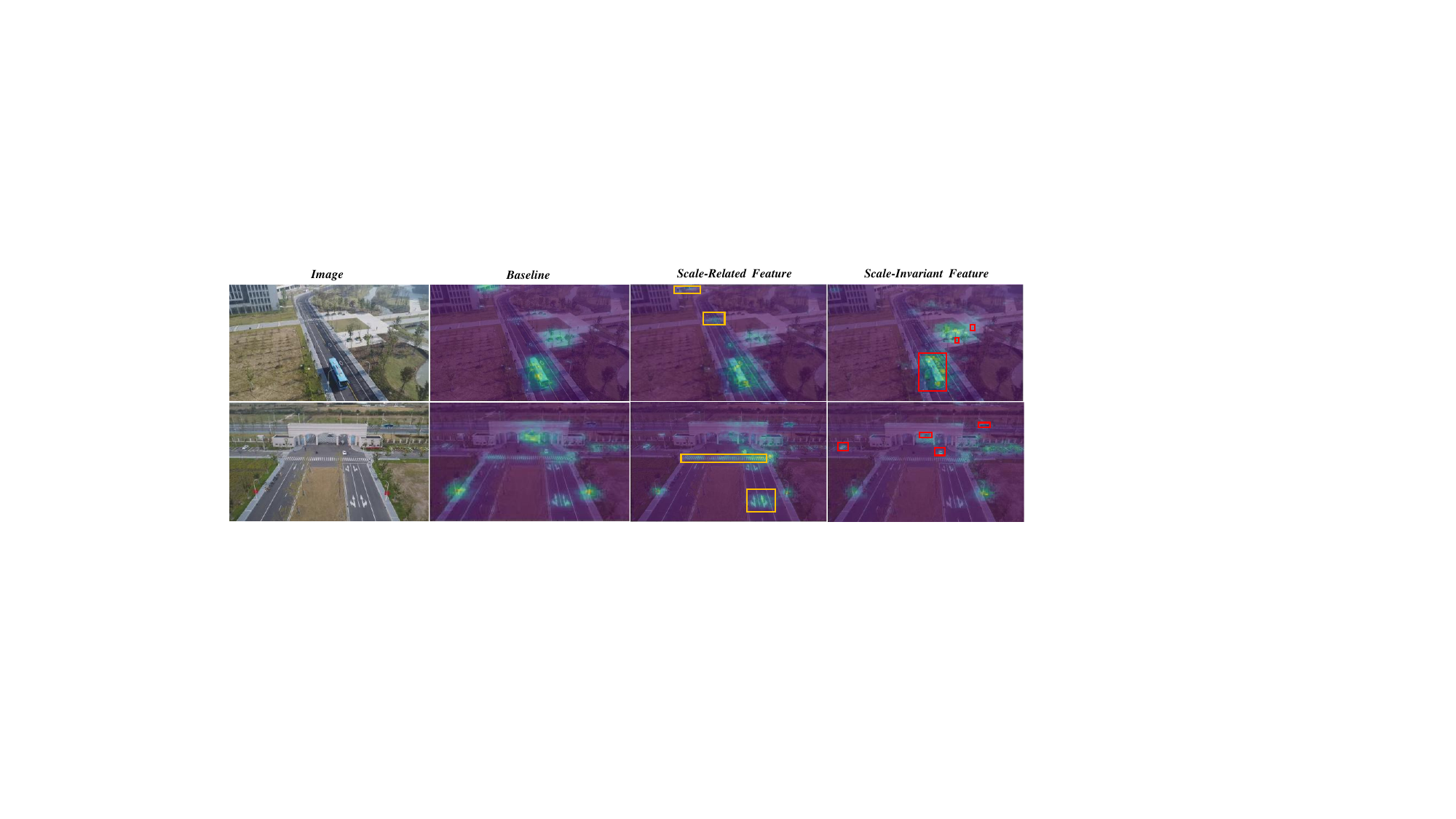}
    \caption{Visualization of scale-related and scale-invariant features. Scale-related features are not only focused on the foreground but also exist in the background (yellow). While scale-invariant features exhibit a stronger concentration on objects (red).
    }
    \label{fig5}
\end{figure*}

\subsection{Visualization Analyses}


We compare the prediction results of our approach and the baseline (vanilla YOLOv7-L) in Fig.~\ref{fig6}. It can be easily observed that the baseline tends to produce a few false or missing detection results (highlighted by orange boxes). Conversely, after employing our method, false and missing detections are suppressed. Specifically, our approach can accurately detect the bus in Fig.~\ref{fig6} (a), cars in (b) and (c), as well as people in (d). This experiment intuitively illustrates the effectiveness of our approach in boosting detection accuracy.

Fig.~\ref{fig5} visualizes the feature disentanglement result. It can be observed that scale-invariant and scale-related features are tangled in the baseline, while applying SIFDAL can visibly disentangle two types of features.
Specifically, scale-related features exist in both foreground and background. For example, they exhibit evident activation on zebra crossings and surface marks, which tend to be inconducive to object detection.
On the contrary, scale-invariant features mainly focus on objects (\eg, cars and buses). The visualization result demonstrates the disentangling effectiveness of our approach and accounts for the reason that scale-invariant features can enhance the model performance.

\section{Conclusions} \label{discussion}
In this paper, we introduced a Scale-invariant Feature Disentanglement via Adversarial Learning (SIFDAL) method to enhance the UAV-based object detection accuracy. 
Specifically, we designed a Scale-Invariant Feature Disentangling (SIFD) module and introduced an Adversarial Feature Learning (AFL) training scheme to obtain discriminative scale-invariant features. 
Our SIFDAL can be employed in any FPN-based object detector and experimental results demonstrated the superiority of our approach.
Furthermore, we constructed a multi-scene and multi-modal UAV-based object detection dataset, State-Air. It was captured in a real-world outdoor setting with a wide variety of scenes and weather conditions. We are committed to further enhancing the scope and scale of State-Air, expanding both the coverage and depth of it.

We provide a supervised disentangling method that effectively utilizes auxiliary information to extract advantageous and redundant features. The former are utilized to benefit target tasks while the latter are removed to reduce misguidance. We believe this paradigm can provide new insights for feature disentangling. In the future, we can explore applying the idea of SIFDAL to the general object detection tasks. By providing labels for horizontal distance (e.g., through depth estimation), SIFDAL can disentangle scale-invariant features in the general tasks to enhance the accuracy of detecting small objects.

Nevertheless, our approach requires UAV flight altitude labels for supervision, which limits its applicability to UAV-OD datasets that lack such data. However, as more detailed and comprehensive UAV-OD datasets become available, this limitation will diminish or even be eliminated. We believe that our multi-modal learning approach is poised to significantly advance the field of UAV-based object detection.

\section*{ACKNOWLEDGMENTS}
This work was supported in part by the National Natural Science Foundation of China under Grant 62372155 and Grant 62302149, in part by the Postgraduate Research and Practice Innovation Program of Jiangsu Province under Grant SJCX24\_0183, in part by the Fundamental Research Funds for the Central Universities under Grant B240201077, in part by the Aeronautical Science Fund under Grant 2022Z071108001, in part by the Qinglan Project of Jiangsu Province, and in part by Changzhou Science and Technology under Project 20231313. 


\bibliographystyle{IEEEtran}
\bibliography{SIFDAL}

\end{document}